\documentclass[10pt,twocolumn,letterpaper]{article}

\usepackage{cvpr} %
\usepackage{times}
\usepackage{epsfig}
\usepackage{graphicx}
\usepackage{amsmath}
\usepackage{amssymb}
\usepackage[ruled,vlined]{algorithm2e}

\usepackage[pagebackref=true,breaklinks=true,letterpaper=true,colorlinks,bookmarks=false]{hyperref}

\usepackage[capitalize]{cleveref}

\crefname{section}{Sec.}{Secs.}
\Crefname{section}{Section}{Sections}

\crefname{figure}{Fig.}{Figs.}
\Crefname{figure}{Figure}{Figures}

\crefname{table}{Tab.}{Tabs.}
\Crefname{table}{Table}{Tables}

\crefname{equation}{Eq.}{Eqs.}
\Crefname{equation}{Equation}{Equations}

\crefname{algocf}{Alg.}{Algs.}
\Crefname{algocf}{Algorithm}{Algorithms}

\usepackage{tikz}
\newcommand{\customyinyang}[1][1]{%
    \begin{tikzpicture}[scale=#1*0.07]
      \draw[line width = #1*0.05mm,transform canvas={yshift=0.02cm}] (0,0) circle (1cm);
      \path[fill=black,transform canvas={yshift=0.02cm}] (90:1cm) arc (90:-90:0.5cm)
                        (0,0)    arc (90:270:0.5cm)
                        (0,-1cm) arc (-90:-270:1cm);

    \end{tikzpicture}}

\newcommand{\blfootnote}[1]{%
  \begingroup
  \renewcommand\thefootnote{}\footnote{#1}%
  \addtocounter{footnote}{-1}%
  \endgroup
}

\begin{document}

\title{DenseFace: Bias Mitigation in Face Recognition \\via Density-Aware Probabilistic Matching}

\author{ Mansur Bultygov\textsuperscript{1,*} \quad Vadim Seliutin\textsuperscript{2,*,\ensuremath{\ddagger}} \quad Dmitry Nekhaev\textsuperscript{1,\ensuremath{\dagger}} \quad Ivan Laptev\textsuperscript{3} \\[0.4em] \textsuperscript{1}VisionLabs \quad \textsuperscript{2}Amazon Web Services \quad \textsuperscript{3}Mohamed bin Zayed University of Artificial Intelligence \\[0.3em] {\tt\small m.bultygov@visionlabs.ai \quad seliutin@amazon.ae } \\ {\tt\small d.nekhaev@visionlabs.ai \quad ivan.laptev@mbzuai.ac.ae } }

\maketitle

\begin{tikzpicture}[remember picture,overlay]
\node[
    anchor=south,
    text width=0.92\paperwidth,
    align=center,
    font=\fontsize{5.5}{6.2}\selectfont
]
at ([yshift=0.10in]current page.south) {
    \textcopyright{} 2026 IEEE. Personal use of this material is permitted.
    Permission from IEEE must be obtained for all other uses, in any current
    or future media, including reprinting/republishing this material for
    advertising or promotional purposes, creating new collective works,
    for resale or redistribution to servers or lists, or reuse of any
    copyrighted component of this work in other works.
};
\end{tikzpicture}

\blfootnote{ \textsuperscript{*}Equal contribution. \quad \textsuperscript{\ensuremath{\dagger}}Corresponding author. \quad \textsuperscript{\ensuremath{\ddagger}}Work done while at VisionLabs. }

\thispagestyle{empty}

\begin{abstract}

Despite steady progress in face recognition, current face recognition models still suffer from significant demographic biases. While approaches for bias mitigation have been proposed, existing methods often impose constraints on the training procedure and result in the degradation of recognition accuracy. To address this issue, we here introduce a method that reduces racial bias in pre-trained face recognition models without compromising their accuracy. To this end, we model face embeddings of each person by von Mises-Fisher (MF) distribution. We next observe the dependency between demographic attributes and the density of MF distributions, and propose DenseFace, a probabilistic face matching procedure that accounts for differences in MF distributions. Our extensive experiments demonstrate DenseFace to consistently reduce racial bias in strong face recognition models varying in network architectures, training datasets and loss functions. Notably, DenseFace preserves recognition accuracy and requires no retraining of the underlying face recognition model. Our work also investigates previously adopted bias measures and makes suggestions.

\end{abstract}
    
\vspace{-0.4cm}
\section{Introduction}
\vspace{-0.1cm}
\label{sec:intro}

Face verification and identification have been significantly improved over the past decade~\cite{he2016deep, deng2019arcface, schroff2015facenet, dan2023transface, shi2019probabilistic}. 
Nevertheless, current approaches still suffer from demographic biases resulting in unbalanced recognition of people with different gender, race and other attributes~\cite{wang2019racial, conti2022mitigating}. 
For example, in crime investigation, biased predictions may lead to unfair results for people across different demographic groups.

\begin{figure}[t]
  \centering

  \includegraphics[width=0.99\linewidth] {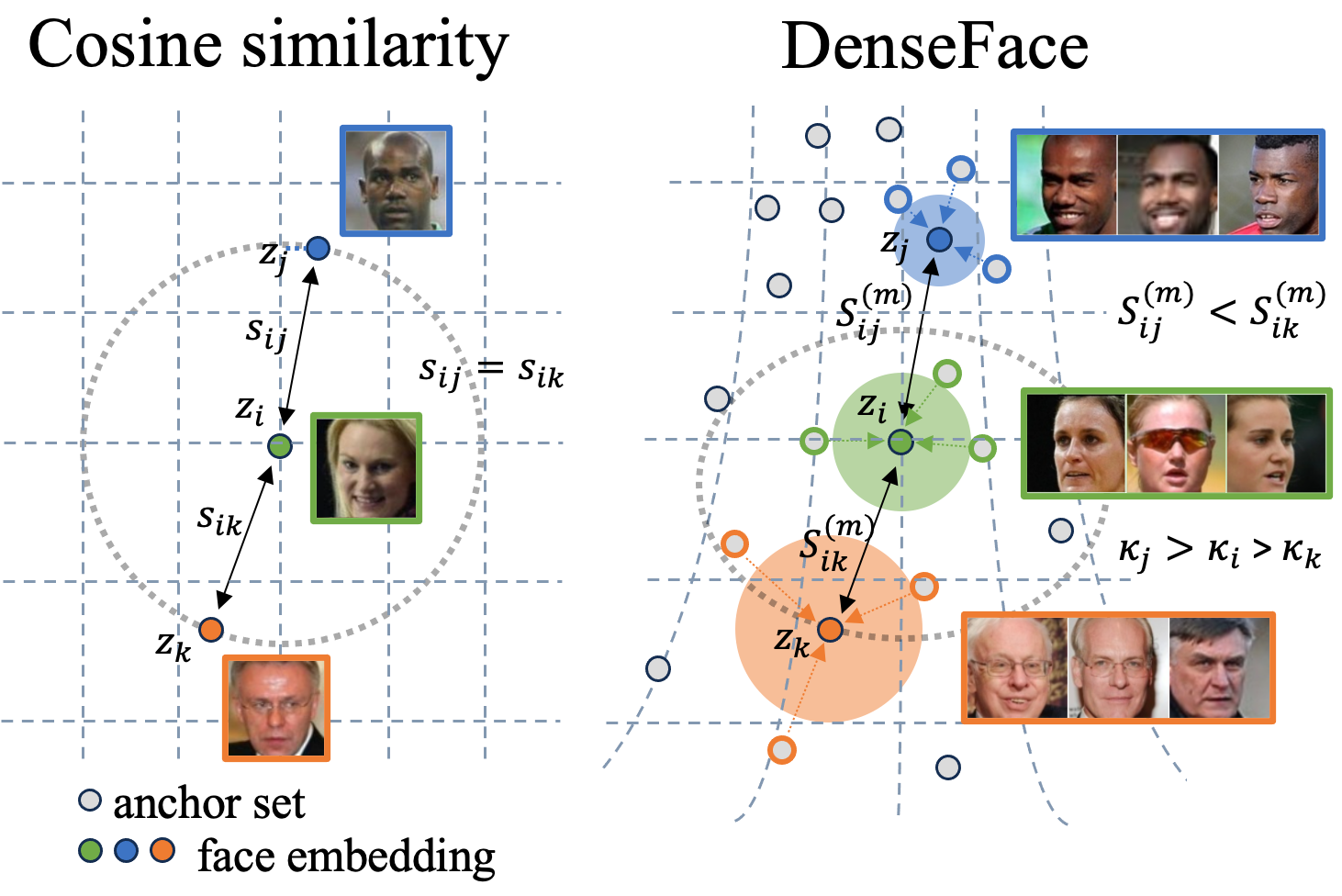}
  \caption{Illustration of density-aware probabilistic matching for bias mitigation. Face embeddings of anchor set identities are used to evaluate local densities $\kappa_{i}$, $\kappa_{j}$,  $\kappa_{k}$, for embeddings of different identities $z_{i}$, $z_{j}$ and $z_{k}$ based on their nearest neighbors. 
  Larger colored circles around face embeddings (right) correspond lower local density. Despite cosine similarities $s_{ij}$ and $s_{ik}$ being the same (left), probabilistic density matching allows to adjust the similarity so that $S^{(m)}_{ik} > S^{(m)}_{ij}$, because embedding $z_j$ is in the region with higher local density than embedding $z_k$.}
  \label{fig:demo}
  \vspace{-0.5cm}
\end{figure}

\begin{figure*}
    \centering
    \includegraphics[width=1.0\textwidth]{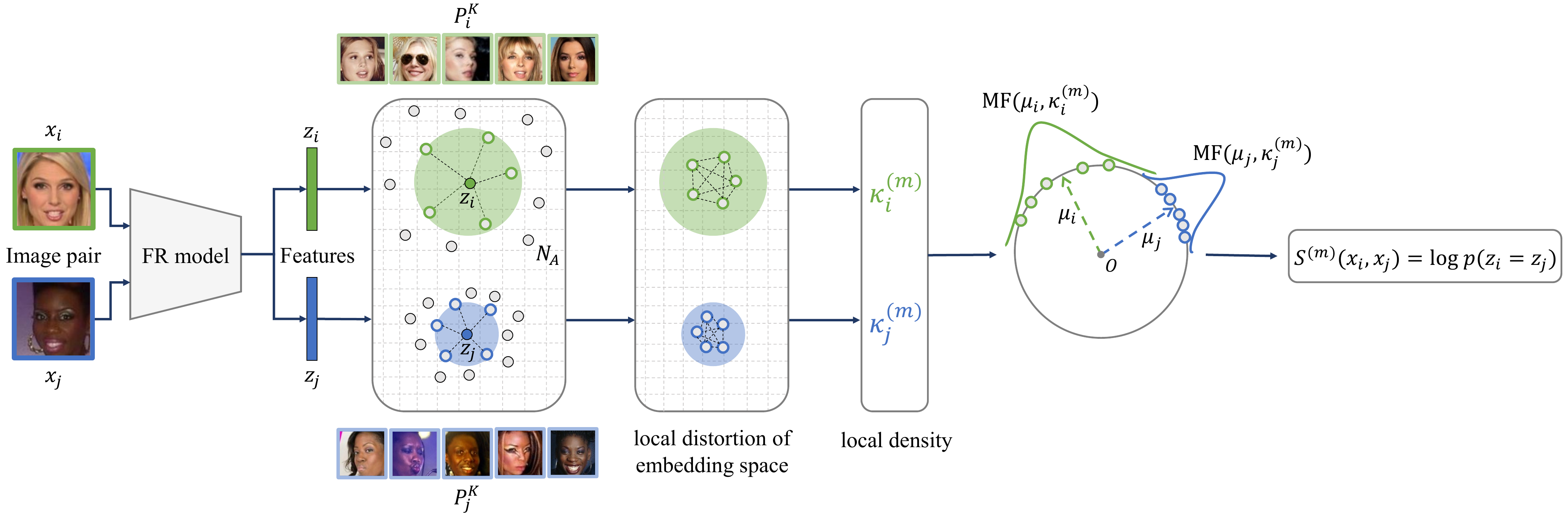}
    \vspace{-0.6cm}
    \caption{Pipeline of density-aware probabilistic matching. Face recognition model takes two input images $x_i$ and $x_j$ for verification task and generates embeddings $z_i$ and $z_j$. $K$ nearest neighbors sets $P^{K}_i$ and $P^{K}_j$ for face embeddings $z_i$ and $z_j$ are obtained from anchor set $N_A$ (Sec. \ref{sec:3.1}).
    Then, we locally distort embedding space by applying margin $m$ in \cref{eq: kappa} to estimate local densities $\kappa^{(m)}_i$ and $\kappa^{(m)}_j$ of the $P^{K}_i$ and $P^{K}_j$ (Sec. \ref{sec:3.2}). Bigger radius of colored circle around the face embedding corresponds to smaller local density. Probabilistic embedding representations $p(z|\mu_i, \kappa^{(m)}_i)$ and $p(z|\mu_j, \kappa^{(m)}_j)$ are then constructed as MF distributions with $\mu_i=z_i$ and $\mu_j=z_j$. Mutual likelihood of these representations to belong to the same identity is considered as matching score for face verification (Sec. \ref{sec:3.3}).
    } 
    \label{fig:pipeline}
    \vspace{-0.2cm}
\end{figure*}

Some of the recent methods for bias mitigation focus on balanced datasets \cite{wang2020mitigating, sevastopolskiy2023boost, gwilliam2021rethinking}.
While training on such datasets can reduce recognition bias, limited sizes of such datasets typically imply degradation of recognition accuracy.
Other methods address bias reduction with alternative network architectures \cite{wang2019racial, gong2021mitigating, huang2023gradient, dooley2024rethinking} and loss functions \cite{wang2020mitigating, xu2021consistent, liu2022learning, ma2023invariant, kotwal2024mitigating}.
Such methods, however, typically require retraining followed by the loss of performance~\cite{conti2022mitigating}.

Following~\cite{terhorst2020post, terhorst2020comparison, dhar2021pass, conti2022mitigating, linghu2024score} in this work 
we focus on developing a post-training calibration approach that can be applied to any pre-trained face recognition network.
Given face embeddings generated by the target model, we consider each embedding as a sample from von Mises-Fisher (MF) distribution that is known to be a conditioned Gaussian distribution on a unit hyper-sphere. We then calculate local density of the face embedding space using an anchor set~\cite{liu2021dam}. We form a subset from the closest embeddings from the anchor set and estimate its local density~\cite{banerjee2005clustering}. 
We observe the dependency between demographic attributes and the density of MF distributions, and propose {\em DenseFace}, a probabilistic face matching procedure that accounts for differences in MF distribution. It allows to adjust similarity scores based on face embedding densities as shown in \cref{fig:demo}. We also introduce a new margin-based method 
to compute local densities of nearly orthogonal embeddings. For efficient deployment, we additionally introduce DenseFace$^\dagger$, which replaces anchor-set search with a lightweight regressor that estimates local density while keeping the face recognition model fixed.

Our work also revisits the question of measuring face recognition bias. We note that classification accuracy commonly used by the RFW protocol~\cite{wang2019racial} may not be suitable in common face recognition scenarios. Moreover, the RFW protocol does not account for cross-racial matching as typically required in real-world scenarios.
To address these and other limitations, we adopt the bias evaluation protocol of NIST~\cite{NIST} and use false match rates (FMRs) at a fixed threshold as the primary bias metric, while also assessing the verification accuracy on multi-racial and cross-racial pairs.

Our extensive experiments demonstrate DenseFace to consistently reduce biases of strong face recognition models varying in network architectures, training datasets and loss functions. Notably, DenseFace preserves recognition accuracy and requires no retraining of the underlying face recognition model.

In summary, we make the following contributions:
\begin{itemize}

    \item We propose to reduce racial bias in face recognition with a new density-aware matching procedure based on von Mises-Fisher (MF) distribution. To estimate inter-class densities of face embeddings, we introduce local distortion of embedding spaces and show the importance of using balanced anchor sets.

    \item {\em DenseFace} mitigates bias in pre-trained face recognition models without retraining the underlying recognition model, while preserving recognition accuracy.

    \item Our experiments demonstrate significant bias reduction for state-of-the-art face recognition models. Following NIST \cite{NIST} evaluation of face verification algorithms, we use false match rate (FMR) at a fixed similarity threshold as a deployment-oriented bias evaluation metric.

\end{itemize}

\begin{table*}[h]
    \vspace{-0.1cm}
    \centering
    \footnotesize
    \caption{Comparison of the verification accuracy (\%) of the
methods on RFW validation set. Upper half belongs to models trained on BUPT-Balancedface. Lower half belongs to models trained on large unbalanced datasets. Asterisk “*” indicates that the results are directly taken from the corresponding paper.}
    \label{tab:rfw_accuracy}
    \begin{tabular}{l|cccc|cc}
        \toprule
        Model & Cauc. & Afr. & Asian & Ind. & Avg $(\uparrow)$ & Std $(\downarrow)$ \\

        \cmidrule(lr){1-7}

        RL-RBN-R34* \cite{wang2020mitigating} & 96.27 &  95.00 & 94.82 & 94.68 & 95.19 & 0.63 \\
    
        DebFace-R34* \cite{gong2020jointly} & 95.95 & 93.67 & 94.33 & 94.78 & 94.68 & 0.83 \\

        DAM-R34* \cite{liu2021dam} & 96.30 & 94.51 & 94.31 & 95.20 & 95.08 & 0.78 \\
        
        ArcFace-R50* \cite{xu2021consistent} & 95.55 & 94.95 & 96.68 & 95.47 & 95.66 & 0.63 \\

        CIFP-R50* \cite{xu2021consistent} & 97.08 & 96.47 & 95.75 & 96.77 & 96.52 & 0.49 \\

        GAC-R50* \cite{gong2021mitigating} & 96.27 & 94.40 & 94.32 & 94.77 & 94.94 & 0.79 \\

        StyleGAN-R50* \cite{sevastopolskiy2023boost} & 96.52 & 95.00 & 93.90 & 94.93 & 95.09 & 0.94 \\

        ArcFace-R100* \cite{xu2021consistent} & 96.43 & 94.98 & 97.37 & 96.17 & 96.24 & 0.85 \\

        CIFP-R100* \cite{xu2021consistent} & 97.03 & 95.65 & 97.60 & 96.82 & 96.78 & 0.71 \\
        
        \cmidrule(lr){1-7}
        
        CosFace-R50-Glink360K \cite{an2022killing} & 98.50 & 98.08 & 99.48 & 98.38 & 98.61 & 0.53 \\

        AdaFace-R50-WebFace4M \cite{kim2022adaface} & 97.52 & 96.68 & 98.57 & 97.53 & 97.57 & 0.67 \\
        
        AdaFace-R100-WebFace4M \cite{kim2022adaface} & 98.42 & 97.80 & 99.47 & 98.12 & 98.45 & 0.63 \\

        AdaFace-R100-WebFace12M \cite{kim2022adaface} & 98.87 & 98.55 & 99.43 & 98.67 & 98.88 & 0.34 \\
        \bottomrule
    \end{tabular}
    \vspace{-0.5cm}
\end{table*}

\vspace{-0.1cm}
\section{Related Work}
\vspace{-0.1cm}
\label{sec:formatting}

\vspace{-0.1cm}
\subsection{Face Recognition}
\vspace{-0.1cm}
Deep learning has dramatically accelerated the progress in face recognition. The emergence of large-scale face datasets \cite{nech2017level, deng2019arcface, an2022killing, zhu2021webface260m} with thousands or millions of identities has allowed for training robust and high precision models. State-of-the-art face encoders are usually based on convolutional \cite{he2016deep, deng2019arcface, schroff2015facenet} or attention \cite{dosovitskiy2020image, dan2023transface} architectures that embed face image into the latent feature space. The choice of a discriminative loss function during training is crucial for an effective faces matching during inference. Having triplet losses \cite{schroff2015facenet} at the dawn of deep face recognition, the clear dominance belongs to softmax-based losses \cite{wang2017normface, liu2017sphereface, wang2018cosface} and their extensions \cite{huang2020curricularface, kim2022adaface, meng2021magface} now. These losses incorporate margins into positive or negative class logits of a training sample to increase inter-class separability and decrease intra-class spread.

\vspace{-0.1cm}
\subsection{Measuring Bias in Face Recognition}
\vspace{-0.1cm}

To measure bias, Wang \etal~\cite{wang2019racial} proposed the Racial faces in-the-wild (RFW) testing dataset. Subsequently, in the number of works \cite{gong2020jointly, sevastopolskiy2023boost, gong2021mitigating, liu2022learning, xu2021consistent, ma2023invariant} model's bias was formulated as the standard deviation of accuracy across four racial subgroups of this test. To complement this, Sevastopolskiy \etal~\cite{sevastopolskiy2023boost} introduced RB-WebFace with the measure of bias as ROC curve values in each subgroup. Liang \etal~\cite{liang2023benchmarking} proposed to benchmark model fairness on synthetic dataset with multiple identities of different races and genders but varying internally in face attributes. Other attempts to measure bias included calculating the geometric mean of FMR and FNMR~\cite{linghu2024score}, the skewed error ratio~\cite{wang2020mitigating, kotwal2024mitigating, huang2023gradient}, the ratio of maximum to minimum values of FMR and FNMR~\cite{conti2022mitigating}, and the trade-off between bias reduction and drop in verification performance~\cite{dhar2021pass}. In this work, we also compare models performance on RFW and RB-WebFace while focusing on FMR values at distance levels fixed across all subgroups. Following NIST \cite{NIST} verification protocol, we argue that this metric better reflects distance dependencies between racial cohorts compared with vanilla accuracy and ROC values.

\vspace{-0.1cm}
\subsection{Bias Mitigation in Face Recognition}
\vspace{-0.1cm}

To address model's bias mitigation from the data perspective, Wang \etal~\cite{wang2020mitigating} introduced BUPT-Balancedface dataset with uniform races distribution and BUPT-Globalface with a distribution close to the world's population. Sevastopolskiy \etal~\cite{sevastopolskiy2023boost} collected unlabeled million-scale African and Asian image sets for self-supervised face encoder pretraining step to ensure equal recognition quality among race groups. Melzi \etal~\cite{melzi2024frcsyn} explored the possibility of bias reduction by introducing the mix of real and synthetic data into training. Zhao \etal~\cite{zhao2025aim} use LLM-guided diffusion to synthesize demography-balanced data and selectively fine-tune bias-sensitive weights (via gradient-difference masks), improving worst-group fairness with minimal accuracy loss. Gwilliam \etal~\cite{gwilliam2021rethinking} demonstrated that balanced training datasets not necessarily reduce bias better than skewed ones. We align with the latter conclusion and show that models trained on large scale unbalanced datasets might possess lesser bias. In addition, collecting an annotated balanced dataset is time consuming and challenging.

From the architectural point of view, Wang \etal~\cite{wang2019racial} introduced information maximization adaptation network to perform domain adaptation from Caucasian group. Gong \etal~\cite{gong2021mitigating} proposed an adaptive layer that learns to avoid discrimination among race and gender groups. Huang \etal~\cite{huang2023gradient} proposed to use adversarial learning on sensitive image regions of each racial domain according to gradient attention maps. Dooley \etal~\cite{dooley2024rethinking} investigated which model-design choices cause the presence of bias and suggested the set of fair backbones found by neural architecture search. In contrast, our method requires neither face encoder modifications nor attribute annotations during inference. DenseFace can be directly applied on top of any trained face recognition model.

Some works focused on ensuring fairness by designing loss functions. Wang \etal~\cite{wang2020mitigating} suggested to learn optimal margins for non-Caucasian groups via reinforcement learning. Xu \etal~\cite{xu2021consistent} improved model fairness through FPR penalty loss function. Liu \etal~\cite{liu2022learning} used sample-level adaptive cosine margins to mitigate the effect of attributes unbalance in training data. Ma \etal~\cite{ma2023invariant} proposed to keep face embedding invariant across demographic attribute groups via self-supervised data partitioning learning strategy. Kotwal \etal~\cite{kotwal2024mitigating} induced score calibration into training by adding regularization loss based on races scores misalignment. Unlike this branch of works, our DenseFace is designed to mitigate bias of a face recognition model without any need for re-training it.

Finally, the problem of acquiring fairness can be addressed with adjusting the decision rule instead of fine-tuning the learned latent space. Terhorst \etal~\cite{terhorst2020post} suggested to assign unique distance thresholds according to face clustering results. In another work ~\cite{terhorst2020comparison}, score distributions of different racial domains were made to be similar by an additional classifier that replaced similarity function. Dhar \etal~\cite{dhar2021pass} introduced an adversarial technique on top of a trained model to suppress sensitive attribute information from face descriptors. Conti \etal~\cite{conti2022mitigating} relied on minimizing von Mises-Fisher loss to project embeddings of a trained model into a fairer latent space. Linghu \etal~\cite{linghu2024score} proposed score normalization methods that reduce ethnicity and gender bias. These approaches demonstrate that demographic bias can also be mitigated through post-hoc calibration. DenseFace instead estimates local inter-class density and incorporates it directly into a probabilistic matching score.

\vspace{-0.1cm}
\section{Methodology}
\vspace{-0.1cm}

Our method is based on the idea that face embedding densities contain demographic information and can be used to adjust matching scores. DenseFace consists of three stages. First, we use a pre-trained face recognition model to represent face embeddings probabilistically. Second, we introduce a local distortion of the embedding space to estimate local inter-class densities. Finally, we match two probabilistic face embeddings using the likelihood that their distributions correspond to the same identity. The whole DenseFace pipeline is shown in \cref{fig:pipeline}.

\vspace{-0.1cm}
\subsection{Probabilistic face embedding representation}\label{sec:3.1}
\vspace{-0.1cm}

Having pre-trained neural network $Z(x)$, we consider face embedding $z_i = Z(x_i)$ of the face image $x_i$ as probabilistic distribution $p(z)$ in the feature space. As state-of-the-art $d$-dimensional face embeddings are usually trained with assumption that during inference they are set on a unit hyper-sphere $\mathbb{S}^{d-1}$ \cite{deng2019arcface}, the MF distribution is a reasonable choice to represent face embedding \cite{banerjee2005clustering, conti2022mitigating, li2021spherical, nech2017level}. The probability density of MF distribution with mean direction $\mu_i \in \mathbb{S}^{d-1} $ and density parameter $\kappa_i> 0$ is defined as follows:

\vspace{-0.1cm}
\begin{equation}
    p(z|\mu_i, \kappa_i) = C_{d}(\kappa_i) e^{\kappa_i \mu_i^{T} z}
    \label{tmp}
\end{equation}

\vspace{-0.3cm}
\begin{equation}
    C_{d}(\kappa_i) = \frac{\kappa_i^{\frac{d}{2}-1}}{(2 \pi)^{d/2} I_{\frac{d}{2}-1}(\kappa_i)}
    \label{eq: cd_kappa}
\end{equation}
\vspace{-0.1cm}
$I_{\frac{d}{2}-1}$ stands for the modified Bessel function of the first kind at order $\frac{d}{2}-1$. 
We follow \cite{banerjee2005clustering} to estimate $\kappa_i$ values:
\vspace{-0.1cm}
\begin{equation}
    \kappa_i = \frac{r_i(d-r_i^2)}{1-r_i^2}
    \label{eq: kappa}
\end{equation}
\vspace{-0.1cm}
\vspace{-0.1cm}
\begin{equation}
    r_i = \frac{\left\Vert\sum_{k=1}^{K}\xi^{k}_i\right\Vert}{K}
    \label{eq: r}
\end{equation}
where $\xi^{i}_k$ are the face embeddings belonging to this distribution and $\Vert \cdot \Vert$ stands for $l_2$-norm. \cref{fig:demo} illustrates estimated density parameters $\kappa_i$, $\kappa_j$ and $\kappa_k$ of the MF distribution for face embeddings - it is higher in regions with high density of face embeddings in the feature space. In practice, we build on DAM~\cite{liu2021dam} and estimate $p(z|\mu_i,\kappa_i)$ using an embedding anchor set $N_A$ obtained from image anchor set $X_A$. DenseFace further uses identity-averaged demographically balanced anchors, local distortion (Sec.~\ref{sec:3.2}), and probabilistic matching (Sec.~\ref{sec:3.3}). We set $\mu_i$ equal to the face embedding $z_i$. For $\kappa_i$ calculation we find $K$ nearest neighbors $\xi^k_i \in N_A, k=1 \dots K$ of the face embedding $z_i$. We refer to this quantity, estimated from the embedding anchor set, as local density. Nearest neighbors embeddings form set $P^K_i \subseteq N_A$.

\begin{figure}[t]
  \centering
  \includegraphics[width=0.85\linewidth] {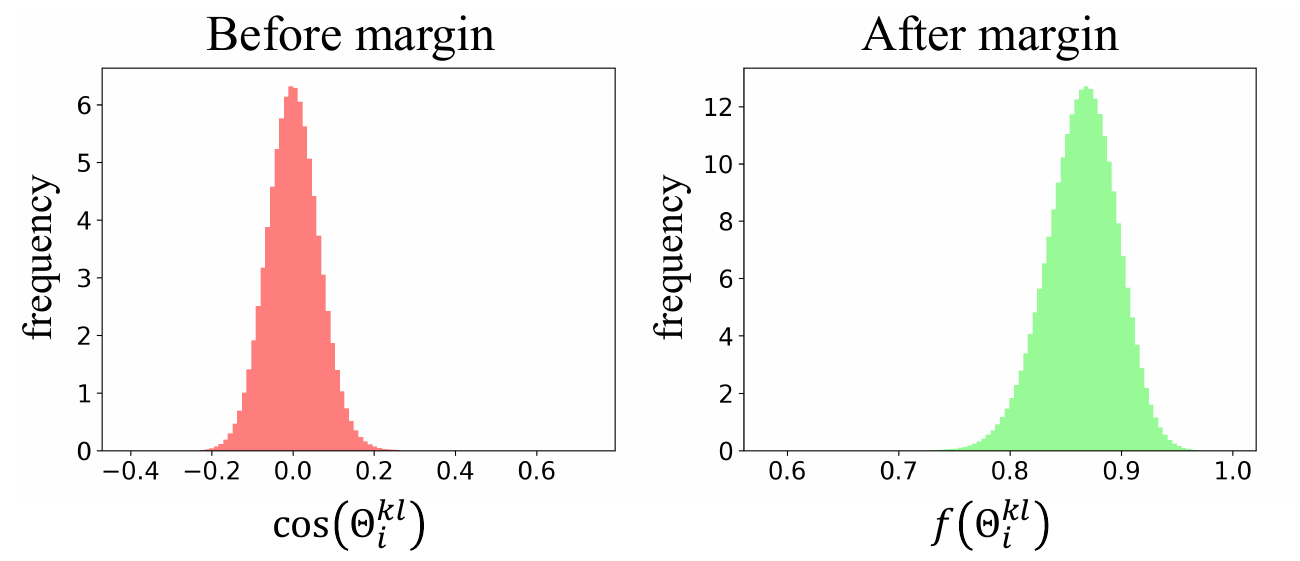}
  \caption{Distributions of pairwise cosine similarities $\cos(\Theta^{kl}_i)$ between anchor set embeddings on the left and $f(\Theta^{kl}_i)$ from \cref{eq: f(theta)} on the right.
  \vspace{-0.52cm}}
  \label{fig: neg_to_pos}
\end{figure}

\vspace{-0.1cm}
\subsection{Local distortion of embedding space}\label{sec:3.2}
\vspace{-0.1cm}

\begin{figure}[t]
  \centering
  \includegraphics[width=0.6\linewidth] {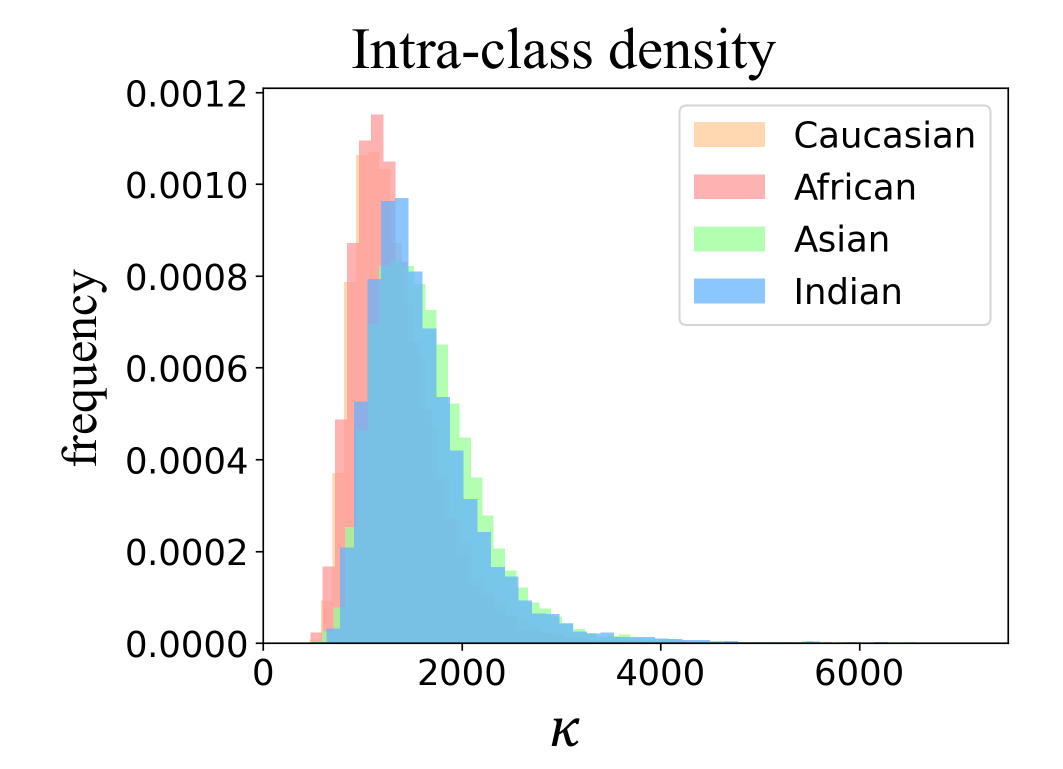}
  \caption{Distributions of intra-class densities $\kappa$ of RB-WebFace embeddings for different racial groups. These distributions are quite similar for all racial groups and do not represent racial bias of face embeddings.} %
  \label{fig: kappa_intra}
  \vspace{-0.4cm}
\end{figure}

Previously, similar approach for face embedding density calculation based on MF distribution was used in \cite{oinar2023kappaface}, where face embeddings of the same identity were used as the embedding anchor set. However, we found that densities computed based on the intra-class distribution of face embeddings do not really represent racial bias (see \cref{fig: kappa_intra}). For example, if we consider class with near similar face images (i.e., obtained from consequential frames of some video) it will have extremely high density no matter which gender or race does the person form this class belong to. Thus, in our method we mainly focus on inter-class face embedding density. We set anchor set embeddings from $N_A$ as mean representation of separate identities. 

\begin{figure}[t]
  \centering
  \includegraphics[width=0.9\linewidth] {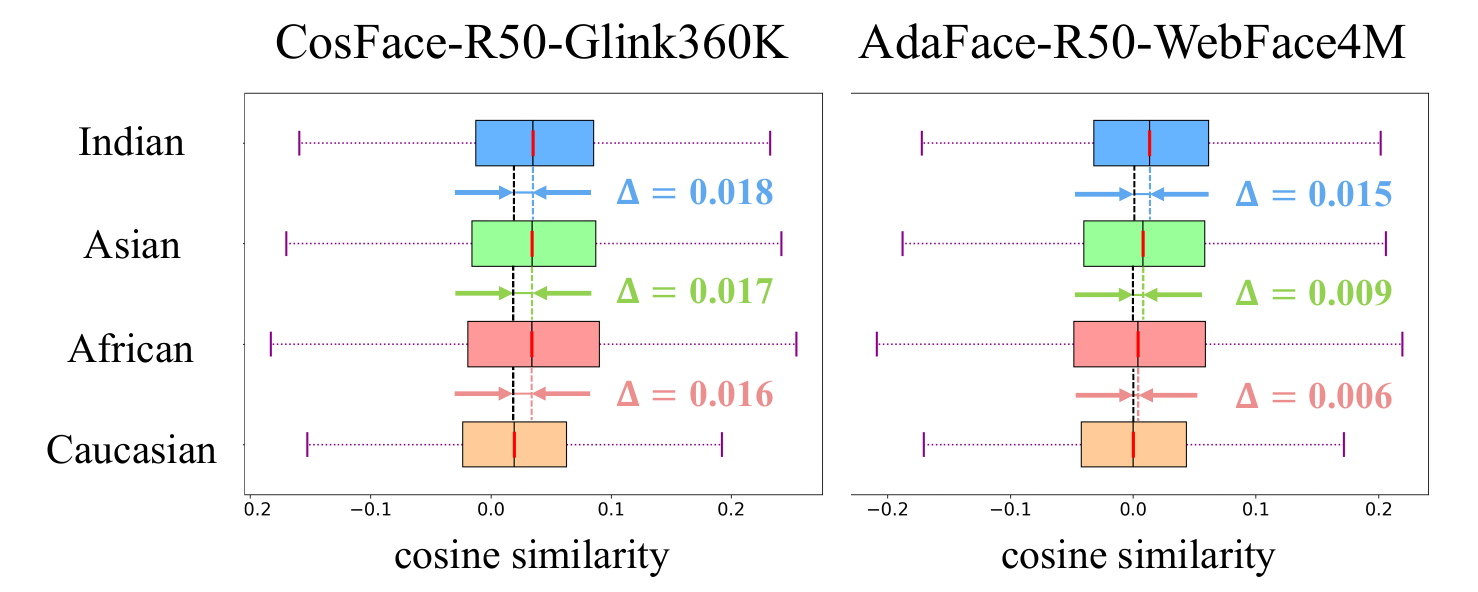}
  \caption{Cosine similarities for negative pairs of RFW test for each racial group. The difference between mean cosine similarities for Caucasian group and other races is denoted as $\Delta$. According to \cref{tab:rfw_accuracy} CosFace-R50-Glink360K has lower racial bias than AdaFace-R50-WebFace4M. However, it has larger $\Delta$ for all racial subgroups which indicates higher racial bias.}
  \label{fig: cos_bias}
  \vspace{-0.4cm}
\end{figure}

\begin{figure}[t]
  \centering
  \includegraphics[width=1.0\linewidth] {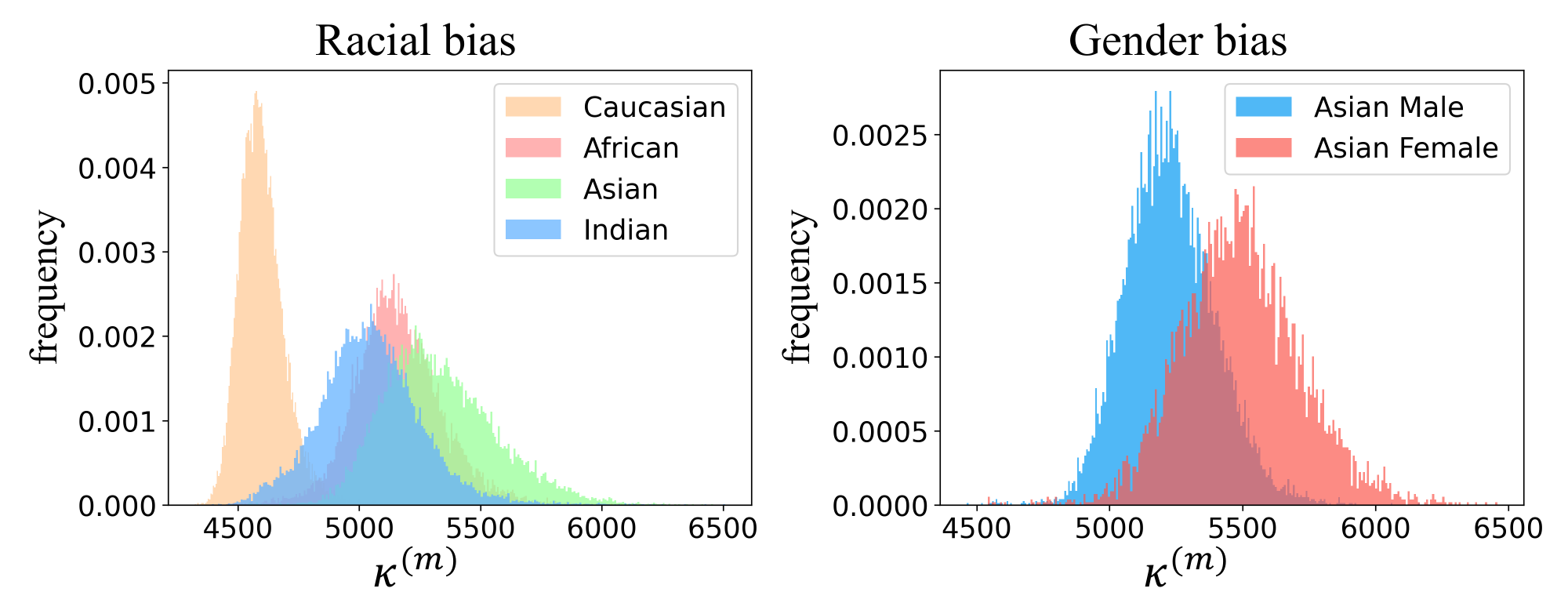}
  \caption{Distributions of inter-class local densities $\kappa^{(m)}$ of RB-WebFace embeddings for different racial groups (on the left) and genders (on the right). These distributions reveal demographic differences in local embedding density.} %
  \label{fig: kappa_bias}
  \vspace{-0.5cm}
\end{figure}

For set of embeddings $\xi^k_i \in P^K_i$ its sum in \cref{eq: r} can be rewritten as
\vspace{-0.6cm}

\begin{equation}
    \left\Vert\sum_{k=1}^K \xi^k_i \right\Vert = \sqrt{\sum_{k=1}^K \Vert \xi^k_i \Vert^2 + 2\sum_{k\neq l} \Vert \xi^k_i\Vert \Vert \xi^l_i \Vert \cos{\Theta^{kl}_{i}}}
    \label{eq: cosines_old}
\end{equation}
where $\cos{\Theta^{kl}_{i}}$ are pairwise cosines between $\xi^k_i, \xi^l_i \in P^K_i$. In practice, when we calculate densities $\kappa_i$, there is a problem of near-similar values of $\kappa_i$ for all face embeddings because of near-orthogonality of all of the embeddings from the anchor set: all $\cos{\Theta^{kl}_{i}}$ are concentrated near zero value and $\left\Vert\sum_{k=1}^K \xi^k_i \right\Vert \simeq \sqrt{\sum_{k=1}^K \Vert \xi^k_i \Vert^2} = \sqrt{K}$ . To fix this issue, we introduce angular margin $m$ inside the cosine function:

\vspace{-0.5cm}
\begin{equation}
    \left\Vert\sum_{k=1}^K \xi^k_i \right\Vert^{(m)} := \sqrt{\sum_{k=1}^K \Vert \xi^k_i \Vert^2 + 2\sum_{k\neq l} \Vert \xi^k_i\Vert \Vert \xi^l_i \Vert f(\Theta^{kl}_{i})}
    \label{eq: cosines_new}
\end{equation}
\vspace{-0.3cm}
\begin{equation}
    f(\Theta^{kl}_{i}) = \begin{cases}
\cos(\Theta^{kl}_{i} - m) & \text{if } \Theta^{kl}_{i} > m \\
1 & \text{otherwise.}
\end{cases}
    \label{eq: f(theta)}
\end{equation}

As a result, the distribution of $\cos\Theta^{kl}_{i}$ shifts to higher values in \cref{fig: neg_to_pos}. Density value computed with margin $m$ is denoted as $\kappa^{(m)}_i$. Distribution of $\kappa^{(m)}_i$ for different racial groups and genders is shown in \cref{fig: kappa_bias}. It can be seen that racial group differences are now reflected in densities: faces of non-Caucasian identities tend to obtain higher densities, which correlates with the fact that they are harder to recognize for state-of-the-art face recognition models. We call the procedure of  margin insertion as local distortion of embedding space, because it acts as a local squeezing operation that increases cosine similarities between anchor embeddings. Local densities $\kappa^{(m)}$ can further be learned to accelerate embedding matching and avoid anchor-set search during inference.

\vspace{-0.1cm}
\subsection{Density-aware probabilistic matching}\label{sec:3.3}
\vspace{-0.1cm}
During matching, we need to calculate similarity score between two face images $x_i$ and $x_j$. For most of the state-of-the-art face recognition models classic approach of cosine similarity $s_{ij} =  \cos(z_i^T z_j)$ between face embeddings $z_i, z_j$ is used. 

For probabilistic embedding matching we use mutual likelihood of two probability distributions $p(z|\mu_i, \kappa_i)$ and $p(z'|\mu_j, \kappa_j)$ to belong to the same person as a matching score:

\vspace{-0.5cm}
\begin{equation}
    \begin{split}
    S_{ij} =  \log \iint \limits_{\mathbb{S}^{d-1} \times \mathbb{S}^{d-1}} p(z|\mu_i, \kappa_i) p(z'|\mu_j, \kappa_j) %
    \delta(z - z') dzdz'
    \label{eq: denseface matching 1}
    \end{split}
\end{equation}

\vspace{-0.3cm}
We follow \cite{li2021spherical} for mutual likelihood estimation of two MF distributions on unit hyper-shere and use $\kappa^{(m)}$ as local densities:

\vspace{-0.6cm}
\begin{equation}
    \begin{split}
    S^{(m)}_{ij} = \log C_d(\kappa_i^{(m)}) + \log C_d(\kappa_j^{(m)}) \\ 
    - \log C_d(\Vert \kappa_i^{(m)} \mu_i + \kappa_j^{(m)} \mu_j \Vert)
    \label{eq: denseface matching 2}
    \end{split}
\end{equation}
\vspace{-0.3cm}

Overall, we call our density-aware probabilistic matching as DenseFace. Final scheme of our method is shown in \cref{fig:pipeline}. \cref{alg: DenseFace} describes whole DenseFace pipeline.

\begin{algorithm}%
\SetAlgoLined
\KwData{Images $x_i$ and $x_j$, image anchor set $X_A$, face recognition network $Z(x)$.}
\KwResult{$S^{(m)}_{ij}$ - similarity score between $x_i$ and $x_j$.}
1. Compute face embeddings:\ 
$z_i = Z(x_i), z_j = Z(x_j)$\

2. Compute embedding anchor set $N_A$:\
$N_A = \{Z(x_m)\mid x_m \in X_A\}$\

3.Compute nearest-neighbor embedding sets $P^K_i$ and $P^K_j$.%

4. Estimate local densities $\kappa^{(m)}_i$ and $\kappa^{(m)}_j$ with \cref{eq: kappa} and \cref{eq: cosines_new}.

5. Compute $S^{(m)}_{ij}$ with \cref{eq: denseface matching 2}.
\caption{DenseFace algorithm}
\label{alg: DenseFace}
\end{algorithm}

\vspace{-0.3cm}
\section{Experiments}
\subsection{Datasets}
\vspace{-0.1cm}
We evaluate our method on RFW ~\cite{wang2019racial} and RB-WebFace ~\cite{sevastopolskiy2023boost} benchmarks. For building anchor image set and training our learning-based approach, we utilize Glint360K ~\cite{an2022killing} dataset. We consider models trained on MS1MV2 ~\cite{deng2019arcface}, Glint360K, WebFace4M, and WebFace12M ~\cite{zhu2021webface260m} datasets.
\vspace{-0.1cm}
\subsection{Implementation Details}
\vspace{-0.1cm}

We follow the works ~\cite{wang2018cosface, deng2019arcface} by aligning and cropping all face images to 112$\times$112 resolution with five landmarks provided by MTCNN ~\cite{zhang2016joint}.

We employ modified versions of ResNet-50 and ResNet-100 ~\cite{deng2019arcface} to extract 512-dimensional face embeddings.

For balanced image anchor set $X_A$ construction, we use pre-trained gender and race classifiers only offline on Glint360K. No demographic classifier or label is required for query images at inference, although classifier errors may affect anchor-set balance. Then, we randomly select 54,000 identities from Glint360K such that the numbers of IDs in subgroups defined by one of the four race groups and gender are equal. We construct embedding anchor set by calculating the average embedding for each identity. The size of the nearest neighbors embedding set $K$ is 128.

For learning-based approach, we train $\kappa^{(m)}$ regression network consisting of two fully-connected layers, ReLU activation functions and batch normalization layers to minimize MSE loss function. The hidden dimension is equal to 256. Models are trained on Glint360K using SGD optimizer with the weight decay of 2e-3 for 100 epochs. Initial learning rate is set to 0.1 and decreased by cosine annealing scheduler. The batch size is set to 2048.

\vspace{-0.1cm}
\subsection{Performance Metrics}
\vspace{-0.1cm}
We argue that common approaches to measure ethnic bias adopted in previous works \cite{gong2020jointly, sevastopolskiy2023boost, gong2021mitigating, liu2022learning, xu2021consistent, ma2023invariant} do not fully characterize model bias in real-world applications.

For instance, the widely used RFW~\cite{wang2019racial} protocol suggests reporting verification accuracy for four ethnicity groups with predefined face image pairs. Optimal distance threshold is defined separately for each race by a k-fold cross validation procedure. A method is said to decrease model's bias if its standard deviation (Std) of accuracy across groups is decreased. However, real biometric systems rarely can afford to set a unique threshold for each racial domain. However, Std alone may be insufficient to characterize deployment behavior. As shown in \cref{tab:rfw_accuracy}, CosFace-R50-Glint360K has lower Std than AdaFace-R50-WebFace4M, while \cref{fig: cos_bias} shows different shifts in impostor-score distributions across racial groups.

RB-WebFace~\cite{sevastopolskiy2023boost} benchmark consists of positive and negative sets of images assembled from WebFace42M dataset. The test includes reporting ROC curves values, namely TPR at FPR = $\{10^{-3}, 10^{-4}\}$, independently for four racial groups. The metrics value is approximated by varying the similarity threshold in a predefined range of values. Although ROC curve consideration eliminates the need to manually set a threshold, RB-WebFace protocol does not reveal that a biased model has different similarity values across ethnicity groups at the same level of FPR, which likewise affects production scenarios.

Following NIST~\cite{NIST}, we evaluate ethnicity bias at a single global threshold set to Caucasian FPR=$10^{-3}$ (analogous to ``MW''). In all FPR-parity tables, FPR is multiplied by $10^3$, so 1 denotes parity with the reference group and deviations in either direction indicate mismatch. In contrast to RB-WebFace protocol, we calculate TPR at FPR having all unique similarity values as thresholds without any predefined range which allows for more precise metric estimation. Instead of using predefined test pairs, we consider all possible pairs of images in RFW. This results in approximately 14 thousand positive and 50 million negative pairs per group compared to 3000 positive and 3000 negative pairs in vanilla RFW. The original pairs of RB-WebFace are preserved.

\vspace{-0.1cm}
\subsection{Debias of SOTA models}
\vspace{-0.1cm}

In this work, we examine strong open-source AdaFace \cite{kim2022adaface} and CosFace \cite{an2022killing} models. These state-of-the-art models were trained on datasets (MS1MV2, Glint360K, WebFace4M, and WebFace12M) with non-uniform ethnicities distribution which is close to real training regimes. Following \cite{dooley2024rethinking}, we question the necessity of preserving a demographic balance in a training dataset. Unlike previous works, we demonstrate that training on large-scale skewed datasets possess lesser or comparable bias according to Std on RFW in comparison to models and debiasing methods trained on BUPT-Balancedface (see \cref{tab:rfw_accuracy}).

To avoid leaks between train and test, we evaluate models trained on WebFace4M and WebFace12M only on RFW. On RB-WebFace, we evaluate models trained on MS1MV2 and Glint360K.

As shown in \cref{tab:rfw_nist} and \cref{tab:rbwf_nist}, applying our DenseFace significantly reduces racial bias for all non-Caucasian domains compared with the cosine similarity baseline, while preserving or improving its verification performance (see \cref{tab:rfw_tpr_fpr} and \cref{tab:rbwf_tpr_fpr}). Additional same-protocol comparisons with post-hoc EM-FAR~\cite{conti2022mitigating} and SN-M3~\cite{linghu2024score} are reported in Supp.~Sec.~\ref{sec:exp_comparison}.

\begin{table}[h]
\centering
\scriptsize
\setlength{\tabcolsep}{3pt}
\caption{Ablation study on matching, RFW, NIST protocol, FPR @ similarity (Caucasian FPR = $10^{-3}$), $\times10^{3}$ scale, the closer to 1 the better. $\overline{\text{DAM}}$ denotes DAM with feature anchor set composed from averaged embeddings of the identity. $\overline{\text{DAM}}~$\customyinyang[1.6] denotes $\overline{\text{DAM}}$ with race and gender balance in anchor set.}
    \label{tab:ablation_matching_rfw_nist}
\begin{tabular}{l|cccc|cccc}
\toprule
Methods & Cauc. & Afr. & Asian & Ind. & Cauc. & Afr. & Asian & Ind. \\
\midrule
& \multicolumn{4}{c|}{CosFace-R50-Glint360K} & \multicolumn{4}{c}{AdaFace-R50-WebFace4M} \\
\midrule
Cosine & 1.00 & 8.82 & 7.69 & 5.96 & 1.00 & 7.67 & 5.84 & 5.62 \\
DAM \cite{liu2021dam} & 1.00 & 8.97 & 7.49 & 6.52 & 1.00 & 7.02 & 5.65 & 6.00 \\
$\overline{\text{DAM}}$ & 1.00 & 8.02 & 6.62 & 6.06 & 1.00 & 6.29 & 4.91 & 5.40 \\
$\overline{\text{DAM}}~$\customyinyang[1.6]  & 1.00 & 0.82 & \textbf{0.67} & 0.56 & 1.00 & 0.72 & 0.57 & 0.56 \\
DenseFace & 1.00 & \textbf{0.91} & 0.43 & \textbf{0.62} & 1.00 & \textbf{1.10} & \textbf{0.63} & \textbf{0.73} \\
\bottomrule
\end{tabular}
\end{table}

\begin{table*}[h]
    \centering
    \footnotesize
    \caption{RFW, NIST protocol, FPR @ similarity (Caucasian FPR = $10^{-3}$), $\times10^{3}$ scale, the closer to 1 the better}
    \label{tab:rfw_nist}
    \begin{tabular}{l|cc|cc|cc|cc}
        \toprule
        & \multicolumn{2}{c|}{CosFace} & \multicolumn{2}{c|}{AdaFace} & \multicolumn{2}{c|}{AdaFace} & \multicolumn{2}{c}{AdaFace} \\
        & \multicolumn{2}{c|}{R50} & \multicolumn{2}{c|}{R50} & \multicolumn{2}{c|}{R100} & \multicolumn{2}{c}{R100} \\
        & \multicolumn{2}{c|}{Glink360K} & \multicolumn{2}{c|}{WebFace4M} & \multicolumn{2}{c|}{WebFace4M} & \multicolumn{2}{c}{WebFace12M} \\
        \cmidrule(lr){2-9}
        & Cosine & DenseFace & Cosine & DenseFace & Cosine & DenseFace & Cosine & DenseFace \\
        \cmidrule(lr){1-9}
        Caucasian & 1.00 & 1.00 & 1.00 & 1.00 & 1.00 & 1.00 & 1.00 & 1.00 \\
        African & 8.82 & \textbf{0.91} & 7.67 & \textbf{1.10} & 7.98 & \textbf{1.22} & 6.66 & \textbf{1.71} \\
        Asian & 7.69 & \textbf{0.43} & 5.84 & \textbf{0.63} & 5.55 & \textbf{0.81} & 4.77 & \textbf{1.19} \\
        Indian & 5.96 & \textbf{0.62} & 5.62 & \textbf{0.73} & 5.93 & \textbf{0.89} & 5.44 & \textbf{1.36} \\
        \bottomrule
    \end{tabular}
\end{table*}

\begin{table*}[h]
    \centering
    \footnotesize
    \caption{RB-WebFace, NIST protocol, FPR @ similarity (Caucasian FPR = $10^{-3}$), $\times10^{3}$ scale, the closer to 1 the better}
    \label{tab:rbwf_nist}
    \begin{tabular}{l|cc|cc|cc}
        \toprule
        & \multicolumn{2}{c|}{AdaFace} & \multicolumn{2}{c|}{AdaFace} & \multicolumn{2}{c}{CosFace} \\
        & \multicolumn{2}{c|}{R50} & \multicolumn{2}{c|}{R100} & \multicolumn{2}{c}{R50} \\
        & \multicolumn{2}{c|}{MS1MV2} & \multicolumn{2}{c|}{MS1MV2} & \multicolumn{2}{c}{Glink360K} \\
        \cmidrule(lr){2-7}
        & Cosine & DenseFace & Cosine & DenseFace & Cosine & DenseFace \\
        \midrule
        Caucasian & 1.00 & 1.00 & 1.00 & 1.00 & 1.00 & 1.00 \\
        African & 2.69 & \textbf{0.80} & 2.86 & \textbf{1.39} & 3.74 & \textbf{0.40} \\
        Asian & 5.16 & \textbf{2.38} & 5.34 & \textbf{3.07} & 9.96 & \textbf{0.71} \\
        Indian & 4.50 & \textbf{1.73} & 4.74 & \textbf{1.75} & 5.19 & \textbf{0.92} \\
        \bottomrule
    \end{tabular}
\end{table*}

\begin{table}[h]
    \centering
    \footnotesize %
    \caption{RFW, TPR @ FPR (\%), AdaFace-R100-WebFace12M}
    \label{tab:rfw_tpr_fpr}
    \begin{tabular}{l|cccc|c}
        \toprule
        Method & Cauc. & Afr. & Asian & Ind. & Avg $(\uparrow)$ \\
        \cmidrule(lr){1-6}
        
        \multicolumn{6}{c}{TPR @ FPR=$10^{-3}$} \\
        \cmidrule(lr){1-6}
        Cosine & 99.82 & 99.62 & 99.41 & 99.51 & \textbf{99.59} \\
        DAM \cite{liu2021dam} & 99.81 & 99.59 & 99.38 & 99.49 & 99.57 \\
        DenseFace & 99.84 & 99.62 & 99.32 & 99.58 & \textbf{99.59} \\
        
        \cmidrule(lr){1-6}

        \multicolumn{6}{c}{TPR @ FPR=$10^{-4}$} \\
        \cmidrule(lr){1-6}
        Cosine & 99.49 & 98.95 & 97.98 & 98.51 & 98.73 \\
        DAM \cite{liu2021dam} & 99.47 & 98.96 & 97.83 & 98.49 & 98.69 \\
        DenseFace & 99.50 & 99.03 & 97.92 & 98.60 & \textbf{98.76} \\
        
        \bottomrule
    \end{tabular}
\end{table}

\begin{table}[h]
    \centering
    \footnotesize %
    \caption{RB-WebFace, TPR @ FPR (\%), AdaFace-R50-MS1MV2}
    \label{tab:rbwf_tpr_fpr}
    \begin{tabular}{lcccc|c}
        \toprule
        Method & Cauc. & Afr. & Asian & Ind. & Avg $(\uparrow)$ \\
        \cmidrule(lr){1-6}
        
        \multicolumn{6}{c}{TPR @ FPR=$10^{-3}$} \\
        \cmidrule(lr){1-6}
        Cosine & 97.54 & 94.63 & 98.36 & 98.41 & 97.24 \\
        DAM \cite{liu2021dam} & 97.63 & 94.80 & 98.32 & 98.39 & 97.28 \\
        DenseFace & 97.57 & 94.94 & 98.20 & 98.52 & \textbf{97.31} \\
        
        \cmidrule(lr){1-6}

        \multicolumn{6}{c}{TPR @ FPR=$10^{-4}$} \\
        \cmidrule(lr){1-6}
        Cosine & 94.78 & 89.62 & 96.06 & 96.04 & 94.12 \\
        DAM \cite{liu2021dam} & 94.92 & 89.91 & 96.16 & 96.02 & 94.25 \\
        DenseFace & 94.83 & 90.10 & 95.89 & 96.27 & \textbf{94.27} \\
        
        \bottomrule
    \end{tabular}
    \vspace{-0.2cm}
\end{table}

\vspace{-0.4cm}
\subsection{Ablation Study}
\vspace{-0.1cm}
In this subsection, we present ablation study of our method. \cref{tab:ablation_matching_rfw_nist} provides a cumulative ablation from DAM~\cite{liu2021dam} to DenseFace. Firstly, we enable anchor set from DAM \cite{liu2021dam} and observe that it does not mitigate bias in terms of FPR @ similarity. Then, we replace embeddings of the anchor set with mean representation of face identities ($\overline{\text{DAM}}$) and slightly reduce bias. Then, we induce racial balance ($\overline{\text{DAM}}~$\customyinyang[1.6]) in image anchor set $X_A$ and mitigate bias drastically. Finally, we apply the DenseFace probabilistic score.

In \cref{tab:ablation_topk_rfw_nist} we perform ablation on the value $K$ of nearest neighbors in \cref{eq: r}. Although for DAM \cite{liu2021dam} $K$ is set to 10, we found that the value of $K$=128 is generally good for bias mitigation with DenseFace.

In \cref{tab:ablation_margin_rfw_nist} we perform ablation on the value of margin $m$ in \cref{eq: cosines_new}. We found that the value $m=\frac{\pi}{3}$ is optimal for each considered model. If margin is too small it is not enough to deal with orthogonality of the embeddings of the anchor set. If it becomes too big it starts to saturate $f(x)$ in \cref{eq: f(theta)} to 1.

\begin{table}[h]
\centering
\setlength{\tabcolsep}{3pt}
\scriptsize
\caption{Ablation study on nearest neighbors number $K$, RFW, NIST protocol, FPR @ similarity (Caucasian FPR = $10^{-3}$), $\times10^{3}$ scale, the closer to 1 the better}
    \label{tab:ablation_topk_rfw_nist}
\begin{tabular}{c|cccc|cccc}
\toprule
$K$ & Cauc. & Afr. & Asian & Ind. & Cauc. & Afr. & Asian & Ind. \\
\midrule
& \multicolumn{4}{c|}{CosFace-R50-Glint360K} & \multicolumn{4}{c}{AdaFace-R50-WebFace4M} \\
\midrule
10 & 1.00 & 0.03 & 0.01 & 0.02 & 1.00 & 0.03 & 0.02 & 0.03 \\
32 & 1.00 & 0.15 & 0.06 & 0.10 & 1.00 & 0.17 & 0.11 & 0.14 \\
64 & 1.00 & 0.39 & 0.16 & 0.27 & 1.00 & 0.48 & 0.28 & 0.34 \\
128 & 1.00 & \textbf{0.91} & 0.43 & 0.62 & 1.00 & \textbf{1.10} & 0.63 & \textbf{0.73} \\
256 & 1.00 & 1.88 & \textbf{1.02} & \textbf{1.24} & 1.00 & 2.10 & \textbf{1.22} & 1.38 \\
512 & 1.00 & 3.29 & 2.01 & 2.05 & 1.00 & 3.33 & 2.06 & 2.26 \\
1024 & 1.00 & 4.89 & 3.26 & 2.95 & 1.00 & 4.54 & 3.03 & 3.20 \\
\bottomrule
\end{tabular}
\vspace{-0.2cm}
\end{table}

\begin{table}[h]
\centering
\setlength{\tabcolsep}{3pt}
\scriptsize
\caption{Ablation study on angular margin, RFW, NIST protocol, FPR @ similarity (Caucasian FPR = $10^{-3}$), $\times10^{3}$ scale, the closer to 1 the better}
    \label{tab:ablation_margin_rfw_nist}
\begin{tabular}{c|cccc|cccc}
\toprule
$m$ & Cauc. & Afr. & Asian & Ind. & Cauc. & Afr. & Asian & Ind. \\
\midrule
& \multicolumn{4}{c|}{CosFace-R50-Glint360K} & \multicolumn{4}{c}{AdaFace-R50-WebFace4M} \\
\midrule
$\pi/4$ & 1.00 & \textbf{1.04} & 2.29 & 1.92 & 1.00 & 2.99 & 2.10 & 2.23 \\
$\pi/3$ & 1.00 & 0.40 & \textbf{0.71} & \textbf{0.92} & 1.00 & \textbf{1.10} & \textbf{0.63} & \textbf{0.73} \\
$5\pi/12$ & 1.00 & 0.06 & 0.04 & 0.20 & 1.00 & 0.30 & 0.21 & 0.25 \\
\bottomrule
\end{tabular}
\vspace{-0.2cm}
\end{table}

\vspace{-0.1cm}
\subsection{Learning-based DenseFace$^\dagger$}
\vspace{-0.1cm}
In this subsection, we present the results of training $\kappa^{(m)}$ regression network for faster matching inference speed. In \cref{tab:learning_based_combined_nist} we provide results for AdaFace-R100-WebFace12M and AdaFace-R50-MS1MV2 on RFW and RB-Webface respectively. DenseFace$^\dagger$ improves bias mitigation over anchor-based DenseFace for both datasets while adding only 0.2\% memory and 1.75\% end-to-end latency over cosine matching (Supp.~Sec.~\ref{sec:computational_cost_analysis}).

\begin{table}[h]
    \centering
    \setlength{\tabcolsep}{3pt}
    \scriptsize
    \caption{Learning-based approach, RFW and RB-WebFace, NIST protocol, FPR @ similarity (Caucasian FPR = $10^{-3}$), $\times10^{3}$ scale, the closer to 1 the better}
    \label{tab:learning_based_combined_nist}
    \begin{tabular}{l|cccc|cccc}
        \toprule

        Methods & Cauc. & Afr. & Asian & Ind. & Cauc. & Afr. & Asian & Ind. \\
        
        \cmidrule(lr){1-9}
        
         & \multicolumn{4}{c|}{RFW} & \multicolumn{4}{c}{RB-WebFace} \\
         & \multicolumn{4}{c|}{AdaFace-R100} & \multicolumn{4}{c}{AdaFace-R50} \\
         & \multicolumn{4}{c|}{WebFace12M} & \multicolumn{4}{c}{MS1MV2} \\
        
        \cmidrule(lr){1-9}
        Cosine         & 1.00 & 6.66 & 4.77 & 5.44 & 1.00 & 2.69 & 5.16 & 4.50 \\
        DenseFace      & 1.00 & 1.71 & 1.19 & 1.36 & 1.00 & \textbf{0.80} & 2.38 & 1.73 \\
        DenseFace$^\dag$ & 1.00 & \textbf{1.38} & \textbf{1.01} & \textbf{1.23} & 1.00 & 0.62 & \textbf{1.87} & \textbf{1.51} \\
        
        \bottomrule
    \end{tabular}
\end{table}

\vspace{-0.2cm}
\subsection{Multi-racial test}
\vspace{-0.1cm}
We also investigated impact of DenseFace on face verification accuracy on multi-racial test that contains all racial groups in one dataset. We combined all images from the RFW dataset into a new test set and performed pairwise comparisons all-vs-all. Multi-racial test contains 40607 images with approximately 57 thousands of positive and 824 millions of negative pairs. Results for the strongest model AdaFace-R100-WebFace12M are presented in \cref{tab:rfw_concat}. DenseFace and DenseFace$^\dag$ consistently improve results over cosine similarity baseline.

\begin{table}[h]
    \centering
    \footnotesize
    \caption{Multi-racial test, RFW, TPR @ FPR (\%)}
    \label{tab:rfw_concat}
    \begin{tabular}{l|cc}
        \toprule

         & TPR @ & TPR @ \\
        Methods & FPR=$10^{-6}$ & FPR=$10^{-5}$ \\

        \cmidrule(lr){1-3}
        \multicolumn{3}{c}{AdaFace-R100-WebFace12M} \\
        \cmidrule(lr){1-3}
        Cosine & 94.72 & 97.84\\
        DenseFace & \textbf{95.29} & \textbf{98.04}\\
        DenseFace$^\dag$ & 95.15 & 97.95 \\
        
        \bottomrule
    \end{tabular}
\end{table}

\vspace{-0.1cm}
\subsection{Quantitative Analysis}
\vspace{-0.1cm}
From \cref{fig: kappa_bias} we make an observation that large values of density $\kappa^{(m)}$ on non-Caucasian domains correspond to high values of FPR for these subgroups (see \cref{tab:rfw_nist} and \cref{tab:rbwf_nist}). This highlights the possibility of using $\kappa^{(m)}$ value as an indicator of racial bias presence.

\begin{figure}[t]
  \centering
  \includegraphics[width=0.95\linewidth] {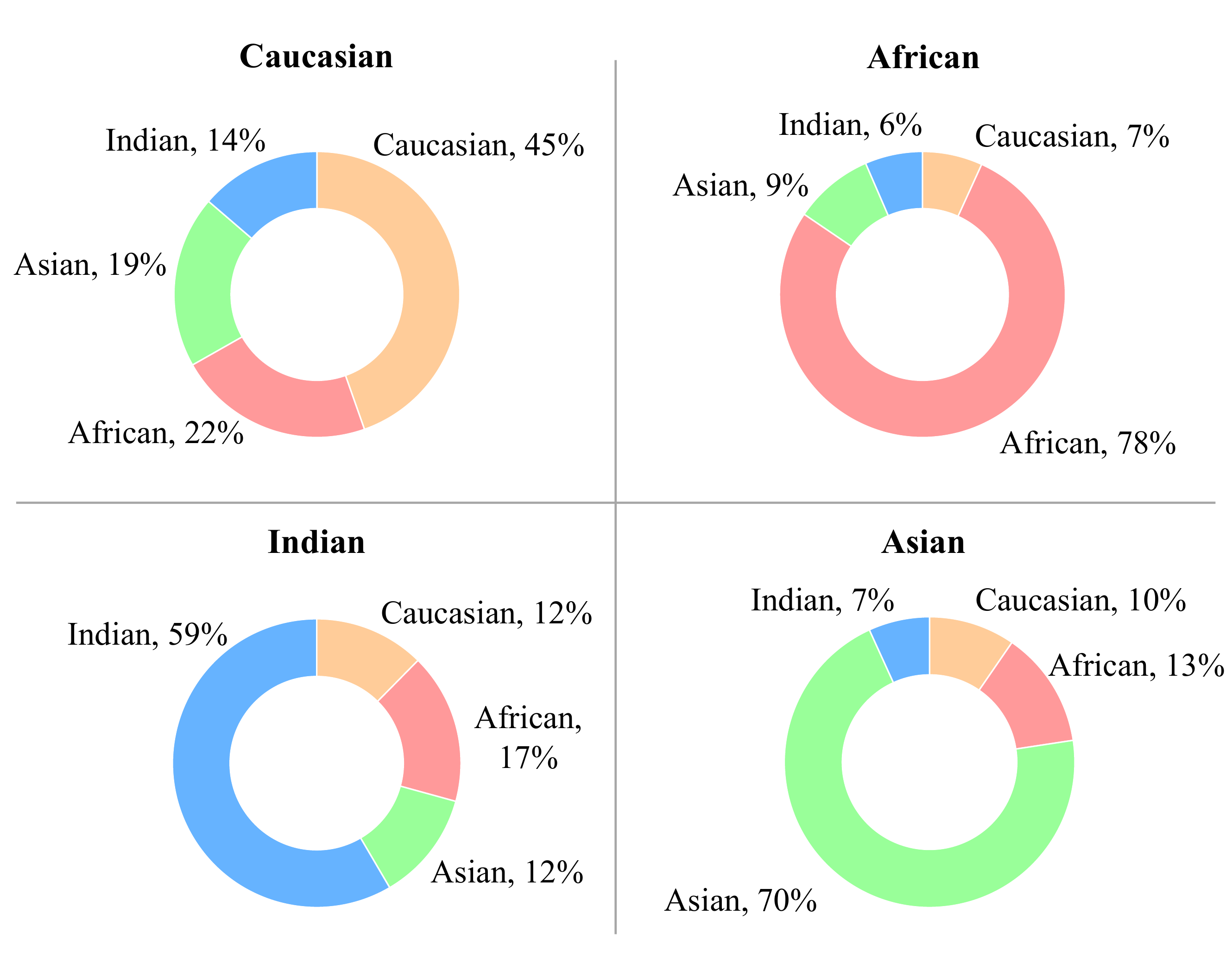}
  \caption{Race distribution among $K$ nearest neighbors from feature anchor set for each race on RFW test. For each racial group the same race represents a larger proportion among nearest neighbors. However, i.e. for Caucasian, 55\% correspond to non-Caucasian races, which indicate that cross-racial interaction is useful for local density estimation.}
  \label{fig: top_k_sectors}
  \vspace{-0.2cm}
\end{figure}

In \cref{fig: top_k_sectors} we provide the distribution of races among $K$ nearest neighbors for each of the RFW racial subgroups. These statistics demonstrate that for each race the majority of neighbors belongs to the same race. This race-correlated neighborhood structure enables DenseFace to estimate local density without ground-truth racial labels at inference. Inducing the demographic balance into the image anchor set allows each race to find the sufficient number of same race neighbors in it. The presence of other races in nearest neighbors distributions can be explained by the cross-racial effect of bias from other attributes like gender or age.

\vspace{-0.2cm}

\vspace{-0.0cm}
\subsection{Qualitative Analysis}
\begin{figure}[t]
  \centering
  \includegraphics[width=1.0\linewidth] {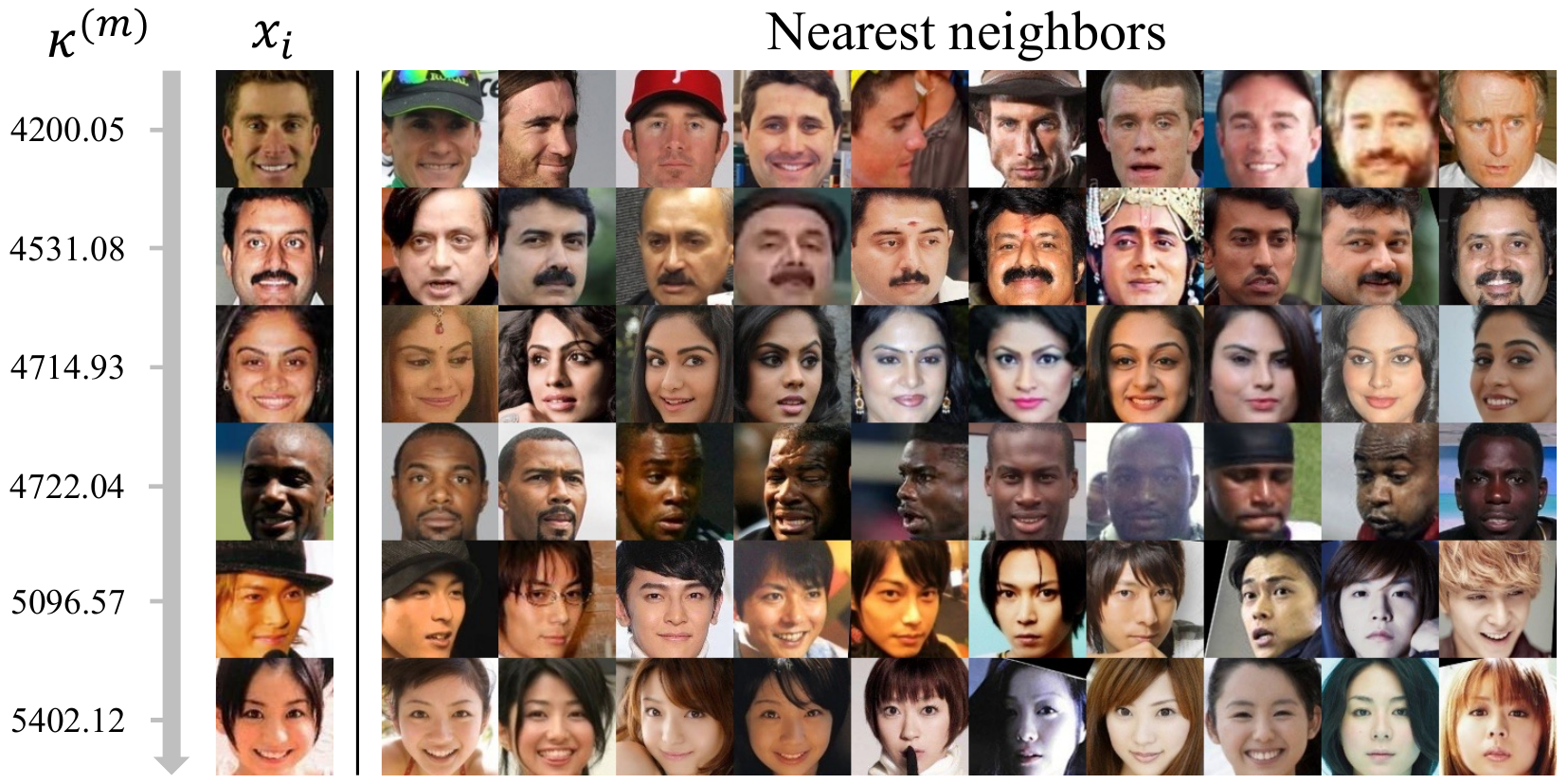}
  \caption{Examples of images with low and high $\kappa^{(m)}$ and their 10 nearest neighbors from the image anchor set. Non-Caucasian and female identities obtain higher values of local densities relatively to Caucasian men identities.}
  \label{fig: kapppa_imgs}
  \vspace{-0.5cm}
\end{figure}

\vspace{-0.1cm}
On the example of face images $x_i$ with low and high $\kappa^{(m)}$ values, we visualize their 10 nearest neighbors from the image anchor set on \cref{fig: kapppa_imgs}. Noticeably, these neighbors not only belong to the same race, but also possess some of the same face attributes like hair style and facial expression. In addition, the ranking of face images $x_i$ according to $\kappa^{(m)}$ aligns with our findings that non-Caucasian and female identities have higher local density values in comparison to Caucasian male group (see \cref{fig: kappa_bias}).

\vspace{-0.1cm}
\section{Conclusion}
\vspace{-0.1cm}
In this paper, we introduced DenseFace, a density-aware probabilistic matching method that mitigates racial bias in pre-trained face recognition models without retraining the underlying recognition model. Experiments across architectures, training datasets, and loss functions show substantial bias reduction while preserving recognition accuracy.

{\small
\bibliographystyle{ieee}
\bibliography{egbib}
}

\clearpage
\setcounter{page}{1}
\twocolumn[
\centering
\Large
\textbf{DenseFace: Bias Mitigation in Face Recognition\\
via Density-Aware Probabilistic Matching}\\
\vspace{0.5em}
Supplementary Material\\
\vspace{1.0em}
]

This Supplementary Material provides additional details for the DenseFace approach. We present training details for BUPT-BalancedFace in Section~\ref{sec:BUPT-BalancedFace}, additional experimental comparisons in Section~\ref{sec:exp_comparison}, evaluation on conventional face benchmarks in Section~\ref{sec:standard_benchmarks}, evaluation results according to RFW protocol in Section~\ref{sec:rfw_results}, limitations of the proposed approach in Section~\ref{sec:computational_cost_analysis} as well as additional details on the anchor set and face image filtering in Sections~\ref{sec:anchorset} and~\ref{sec:identityfilter} respectively.

\section{Training on BUPT-BalancedFace}
\label{sec:BUPT-BalancedFace}

Following common de-biasing research setup, we also train ResNet-50 on BUPT-Balancedface \cite{wang2020mitigating}. Results are presented in \cref{tab:rfw_nist_bupt_training}. Despite training on the balanced dataset, ResNet-50 model trained with CosFace is highly biased. DenseFace mitigates bias for African and Indian domains: it reduces FPR at the fixed threshold approximately fourfold, from 0.00615 to 0.00143 and from 0.00401 to 0.00103, respectively.

\begin{table}[h]
    \centering
    \caption{RFW, training on BUPT-Balancedface, NIST protocol, FPR @ similarity (Caucasian FPR = $10^{-3}$), $\times10^{3}$ scale, the closer to 1 the better}
    \label{tab:rfw_nist_bupt_training}
    \begin{tabular}{l|cccc}
        \toprule

        Methods & Cauc. & Afr. & Asian & Ind. \\

        \cmidrule(lr){1-5}
        \multicolumn{5}{c}{CosFace-R50-BUPT-Balancedface} \\ %
        \cmidrule(lr){1-5}
        Cosine & 1.00 & 6.15 & 1.66 & 4.01 \\
        DenseFace & 1.00 & \textbf{1.43} & \textbf{0.88} & \textbf{1.03} \\
        
        \bottomrule
    \end{tabular}
\end{table}

\section{Comparison with other de-biasing methods}
\label{sec:exp_comparison}

To demonstrate remaining racial bias in models trained with state-of-the-art de-biasing approaches, we evaluated available open-source de-biased models on RFW benchmark with FPR at similarity metric. Although ResNet-34 model trained with CIFP \cite{xu2021consistent} approach on BUPT-Balancedface has less bias than CosFace-R34-Glint360K, it still has huge difference of FPR among different racial groups (\cref{tab:rfw_cifp}). However, applying of DenseFace allows to mitigate remaining bias. 

\begin{table}[h]
    \centering
    \setlength{\tabcolsep}{3pt}
    \footnotesize
    \caption{RFW, NIST protocol, FPR @ similarity (Caucasian FPR = $10^{-3}$), $\times10^{3}$ scale, the closer to 1 the better}
    \label{tab:rfw_cifp}
    \begin{tabular}{l|cccc|cccc}
        \toprule

        Methods & Cauc. & Afr. & Asian & Ind. & Cauc. & Afr. & Asian & Ind. \\
        
        \cmidrule(lr){1-9}

         & \multicolumn{4}{c|}{CIFP-R34 \cite{xu2021consistent}} & \multicolumn{4}{c}{ CosFace-R34-Glint360K} \\
        
        \cmidrule(lr){1-9}
        Cosine         & 1.00 & 5.43 & 1.54 & 3.21 & 1.00 & 9.35 & 7.58 & 5.67 \\
        DenseFace      & 1.00 & \textbf{1.70} & \textbf{0.81} & \textbf{0.80} & 1.00 & \textbf{1.10} & \textbf{0.46} & \textbf{0.73} \\
        
        \bottomrule
    \end{tabular}
\end{table}

We also acknowledge the work of \cite{gong2020jointly, gong2021mitigating, huang2023gradient, liu2022learning, sevastopolskiy2023boost, wang2020mitigating}, but do not compare with them directly as the code, the pretrained models, and the results for our experimental setup were not available for these methods. We reimplemented \cite{conti2022mitigating, linghu2024score} and evaluated these methods in the AdaFace-R100-WebFace12M setup of \cref{tab:learning_based_combined_nist}. The results reported in \cref{tab:comparison_w_post_training} indicate significant improvements of our method.

\begin{table}[h]
    \centering
    \setlength{\tabcolsep}{3pt}
    \caption{RB-WebFace, NIST protocol, FPR @ similarity (Caucasian FPR = $10^{-3}$), $\times10^{3}$ scale, the closer to 1 the better}
    \label{tab:comparison_w_post_training}
    \begin{tabular}{l|cccc}
        \toprule

        Methods & Cauc. & Afr. & Asian & Ind. \\

        \cmidrule(lr){1-5}

        \multicolumn{5}{c}{AdaFace-R100-WebFace12M} \\ %
        
        \cmidrule(lr){1-5}
        EM-FAR \cite{conti2022mitigating} & 1.00 & 6.56 & 4.80 & 6.15 \\
        SN-M3 \cite{linghu2024score} & 1.00 & 4.03 & 2.07 & 2.70 \\
        Cosine         & 1.00 & 6.66 & 4.77 & 5.44  \\
        DenseFace      & 1.00 & 1.71 & 1.19 & 1.36 \\
        DenseFace$^\dag$ & 1.00 & \textbf{1.38} & \textbf{1.01} & \textbf{1.23} \\
        
        \bottomrule
    \end{tabular}
\end{table}

\section{Evaluation on Standard Face Recognition Benchmarks}
\label{sec:standard_benchmarks}

We have performed evaluations for the strongest AdaFace-R100-WebFace12M model on standard benchmarks LFW \cite{huang2008labeled}, CFP-FP \cite{sengupta2016frontal}, AgeDB \cite{moschoglou2017agedb}, CPLFW \cite{zheng2018cross}, CALFW \cite{zheng2017cross}. Along with \cref{tab:rfw_tpr_fpr} and \cref{tab:rbwf_tpr_fpr}, these results (\cref{tab:lfw_cfp_agedb_cplfw_calfw}) demonstrate that DenseFace allows to maintain original accuracy on standard benchmarks while significantly improving the baseline for the racially representative test set (see \cref{tab:rfw_concat}).

\begin{table}[h]
    \centering
    \small
    \setlength{\tabcolsep}{2pt}
    \caption{Verification accuracy (\%) for DenseFace on LFW, CFP-FP, AgeDB, CPLFW, CALFW}
    \label{tab:lfw_cfp_agedb_cplfw_calfw}
    \begin{tabular}{l|ccccc|c}
        \toprule

        Methods & LFW & CFP-FP & AgeDB & CPLFW & CALFW & Avg \\

        \cmidrule(lr){1-7}
        \multicolumn{7}{c}{ AdaFace-R100-WebFace12M} \\ %
        \cmidrule(lr){1-7}
        Cosine & 99.78 & \textbf{99.13} & 98.02 & 94.30 & 95.97& 97.44 \\
        DenseFace & 99.78 & 99.01 & 98.08 & \textbf{94.32} & \textbf{96.10} & \textbf{97.46} \\
        DenseFace$^\dag$ & \textbf{99.83} & 98.96 & \textbf{98.12} & 94.23 & 96.03 & 97.43 \\
        
        \bottomrule
    \end{tabular}
\end{table}

\section{DenseFace on original RFW protocol}
\label{sec:rfw_results}

We also evaluate DenseFace on the standard RFW protocol. The results of state-of-the-art models trained on Glint360K and the subsets of WebFace42M are shown in \cref{tab:denseface_frw_accuracy}. DenseFace approach slightly reduces bias in terms of the standard deviation of the accuracy metric, without reduction of the average accuracy.

\begin{table}[h]
    \centering
    \scriptsize
    \setlength{\tabcolsep}{1.7pt}
    \caption{Verification accuracy (\%) for DenseFace on RFW validation set. R50 is trained with CosFace, R100 is trained with AdaFace}
    \label{tab:denseface_frw_accuracy}
    \begin{tabular}{ll|cccc|cc}
        \toprule
        Model & Matching & Cauc. & Afr. & Asian & Ind. & Avg $(\uparrow)$ & Std $(\downarrow)$ \\

        \cmidrule(lr){1-8}

        R50-Glink360K \cite{an2022killing} & Cosine & 98.50 & 98.08 & 99.48 & 98.38 & 98.61 & 0.53 \\
        \ & DenseFace & 98.55 & 98.33 & 99.43 & 98.45 & 98.68 & 0.44 
        \\

        \cmidrule(lr){1-8}

        R100-WebFace4M \cite{kim2022adaface} & Cosine & 98.42 & 97.80 & 99.47 & 98.12 & 98.45 & 0.63 \\
        \  & DenseFace & 98.55 & 97.78 & 99.43 & 98.13 & 98.48 & 0.62 \\

        \cmidrule(lr){1-8}

        R100-WebFace12M \cite{kim2022adaface} & Cosine & 98.87 & 98.55 & 99.43 & 98.67 & 98.88 & 0.34 \\
        \  & DenseFace & 99.02 & 98.53 & 99.35 & 98.65 & 98.89 & 0.32 \\
        
        \bottomrule
    \end{tabular}
\end{table}

\section{Computational Cost Analysis and Limitations}
\label{sec:computational_cost_analysis}
We have evaluated the computational cost of local density computation with and without the use of the anchor set and compared the results with the Cosine baseline. We use an NVIDIA A100 GPU with batch size 8. As shown in \cref{tab:density_cost}, the proposed regression network in DenseFace$^\dag$  adds negligible memory and latency overhead compared to the baseline.

\begin{table}[h]
    \centering
    \small
    \vspace{-0.1cm}
    \caption{Computational cost in terms of latency and memory. Cosine implies backbone inference to obtain embeddings. DenseFace and  DenseFace$^\dag$ is for the inference and local density estimation based on anchor set or regression network, respectively.}
    \label{tab:density_cost}
    \begin{tabular}{l|cc}
        \toprule

        Methods & Memory, MB & Latency, ms \\

        \cmidrule(lr){1-3}
        \cmidrule(lr){1-3}
        Cosine & 248.52 & 17.71  \\
        DenseFace & 355.23 (+ 42.93\%) & 102.63 (+ 479.54\%) \\
        DenseFace$^\dag$ & 249.04 (+ 0.2\%) & 18.02 (+ 1.75\%)\\
        
        \bottomrule
    \end{tabular}
\end{table}

We also measured the matching speed according to~\cref{eq: denseface matching 2} (see \cref{tab:matching_speed}). We emphasize that the first two terms in~\cref{eq: denseface matching 2} could be computed during the local density calculation phase along with the extraction of face embeddings. Also, operations in the third term such as the weighted sum of two embeddings, taking its norm, and calculating the modified Bessel function of the first kind may incur non-negligible computational cost. To minimize the cost caused by norm operation, we appeal to Numba library \cite{lam2015numba} that allows to translate Python code into highly optimized machine code at runtime. We use look up table (LUT) data structure to replace the modified Bessel function of the first kind to proceed with the cost minimization, 
which results in 40 MB memory trade off to store kappa values and their corresponding function values obtained on RFW and RB-WebFace datasets.
The introduced techniques affect neither bias nor verification performance.
For time cost measurement, we use an AMD EPYC 7702 64-Core CPU.

\begin{table}[h]
    \centering
    \small
    \setlength{\tabcolsep}{2pt}
    \vspace{-0.1cm}
    \caption{Cosine and DenseFace matching time cost. Each method is tested on 10k iterations of randomly generated unit vectors, resulting in the average and standard deviation time.}
    \label{tab:matching_speed}
    \begin{tabular}{l|ccc}
        \toprule

        Methods & Avg. time ratio & Avg. ms & Std. ms \\

        \cmidrule(lr){1-4}
        \cmidrule(lr){1-4}
        Cosine & - &  0.029 & 0.00346 \\
        DenseFace$^\dag$ & 2.5x & 0.071 & 0.00319 \\
        DenseFace$^\dag$(LUT) & 2.1x & 0.061 & 0.00330 \\
        DenseFace$^\dag$(LUT + Numba) & 1.5x & 0.042 & 0.00213\\
        
        \bottomrule
    \end{tabular}
\end{table}

\section{Anchor set identity overlap}
\label{sec:anchorset}

We noticed that there could be an overlap between identities from anchor set and regressor DenseFace$^\dag$ training set. It could result in higher values of $\kappa^{(m)}$ density for such overlapping identities. To prevent this effect, we detected these cases by cosine similarity threshold $0.6$ and excluded these identity duplicates from the nearest neighbor sets $P_i^K$.

\section{Local density as image filter}
\label{sec:identityfilter}

\begin{figure}[h]
    \vspace{-0.1cm}
    \centering
    \includegraphics[width=0.99\linewidth]{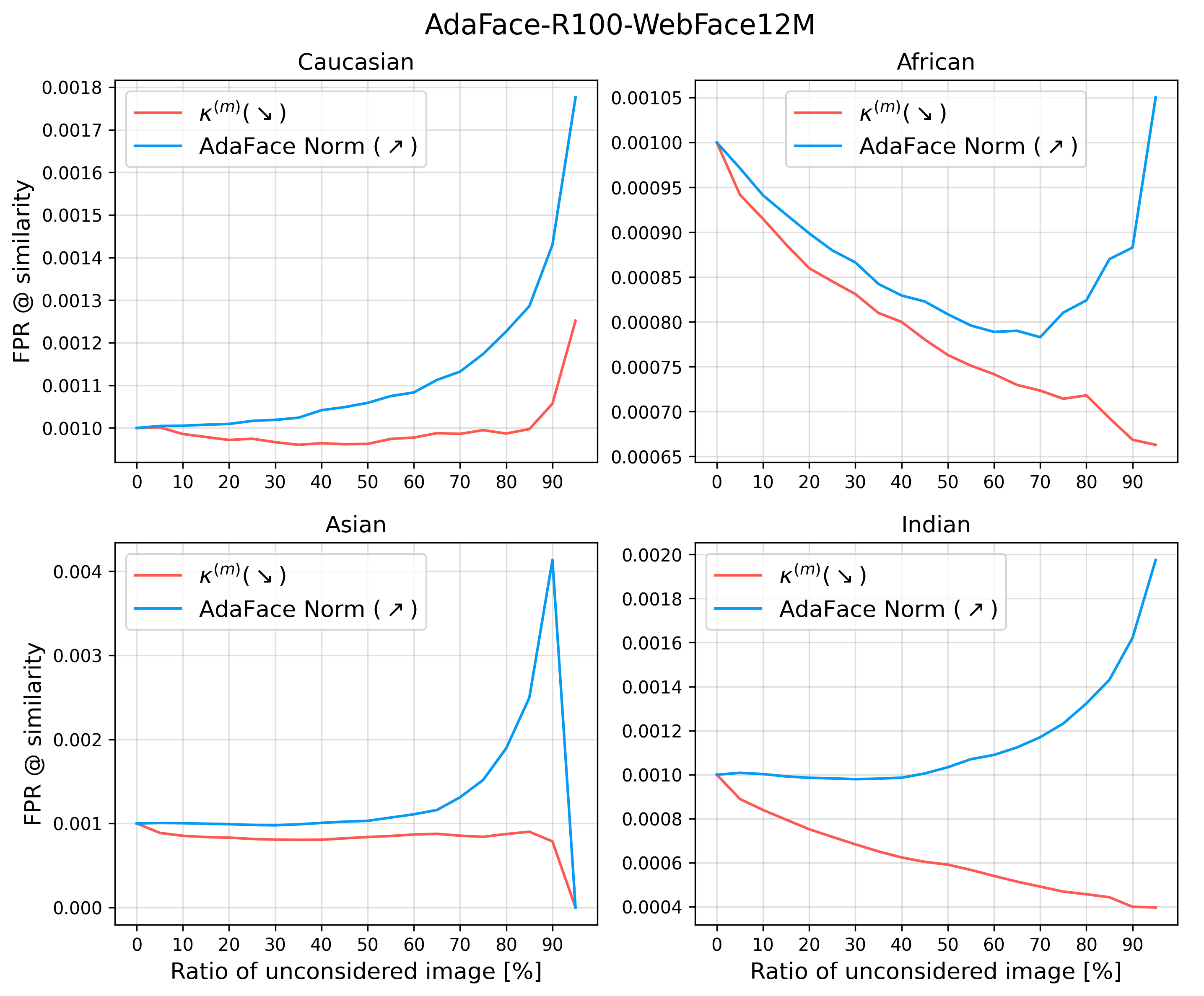}
  \caption{ERC comparison between $\kappa^{(m)}$ density and AdaFace embedding norm as image filter on RFW. The plots show the effect of removing samples with high local density (DenseFace) or low face embedding norm (AdaFace).}
    \label{fig:sup-rfw-kappa-filter-r1000wf12m}
\end{figure}

As face embedding local density helps to mitigate bias in our DenseFace approach, we tried to use $\kappa^{(m)}$ as image filter that allows to remove some samples from the verification task. Since high values of local density indicate high similarity between different identities in this region, we sorted $\kappa^{(m)}$ values in descending order for all test images and removed samples with the highest density values. We varied the percentage of filtered images and calculated FPR at threshold value that initially corresponded to FPR=0.001 and plotted the Error vs. Reject Curves (ERC). The results are shown in \cref{fig:sup-rfw-kappa-filter-r1000wf12m}. Removal of the samples with high local density helps to reduce FPR for African and Indian groups. Caucasian and Asian false positive errors appear to be less sensitive to estimated local densities.

\begin{figure}[h]
    \vspace{-0.1cm}
    \centering
    \includegraphics[width=0.99\linewidth] {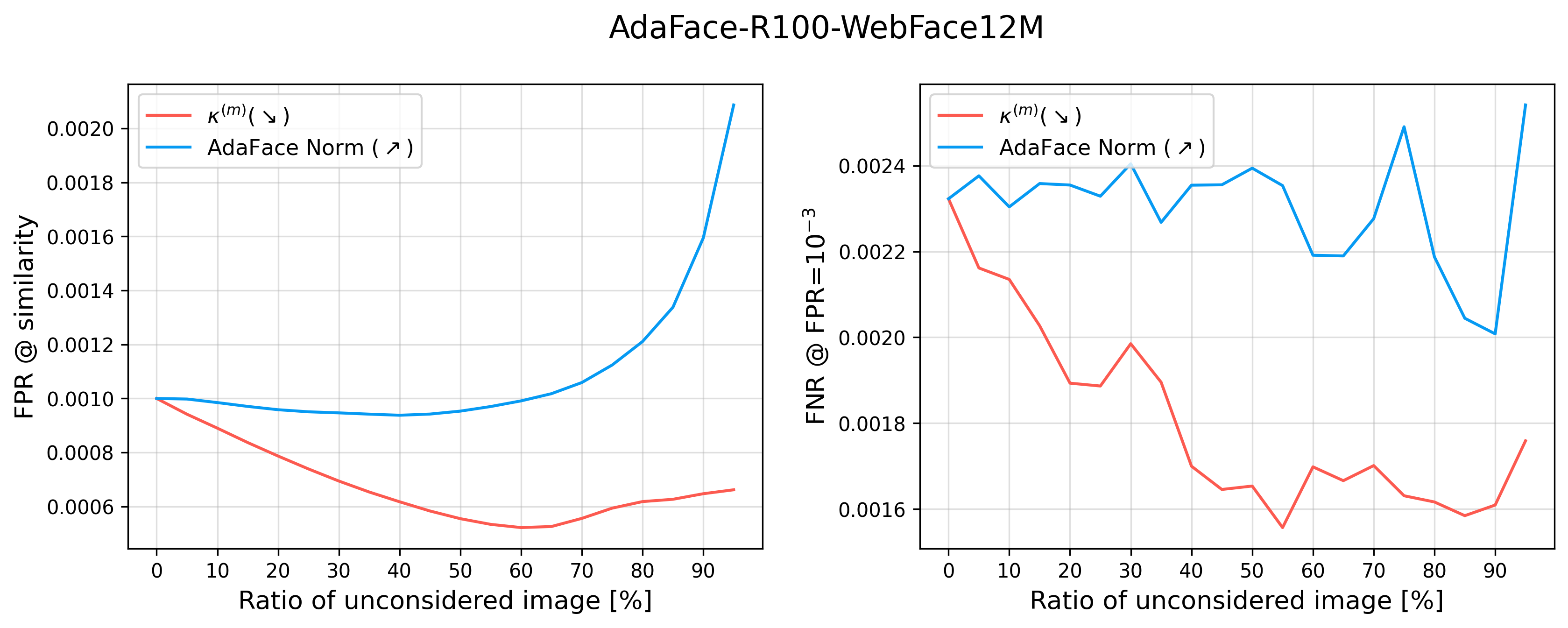}
  \caption{ERC comparison between $\kappa^{(m)}$ density and AdaFace embedding norm as image filter on Multi-racial test, RFW. The plots show the effect of removing samples with high local density (DenseFace) or low face embedding norm (AdaFace).}
    \label{fig:sup-filtration_fpr_at_sim_and_fnr_at_fpr_rfw_concat_adaface_ir101_webface12m}
\end{figure}

We also show that it can be useful to remove samples with the highest density on multi-racial test. In \cref{fig:sup-filtration_fpr_at_sim_and_fnr_at_fpr_rfw_concat_adaface_ir101_webface12m} we plot ERC curves for FPR at fixed initial threshold (FPR=0.001) and FNR (FNR=1-TPR) at fixed level of FPR = 0.001. So, in order to reduce face verification error, there could be two different approaches of sample filtering: we can reduce FPR at the fixed threshold value, or, we can reduce FNR at the fixed FPR value.

\end{document}